# MISSING DATA IMPUTATION FOR SUPERVISED LEARNING[†]

Jason Poulos[‡] and Rafael Valle

*University of California, Berkeley*

*Abstract:* Missing data imputation can help improve the performance of prediction models in situations where missing data hide useful information. This paper compares methods for imputing missing categorical data for supervised classification tasks. We experiment on two machine learning benchmark datasets with missing categorical data, comparing classifiers trained on non-imputed (i.e., one-hot encoded) or imputed data with different levels of additional missing-data perturbation. We show imputation methods can increase predictive accuracy in the presence of missing-data perturbation, which can actually improve prediction accuracy by regularizing the classifier. We achieve the state-of-the-art on the Adult dataset with missing-data perturbation and $k$-nearest-neighbors ($k$-NN) imputation.

*Keywords:* neural networks; decision trees; imputation methods; missing data; perturbation; random forests

[†]The code used for this project is available at https://github.com/rafaelvalle/MDI. The online Supplementary Material is available at http://jvpoulos.github.io/papers/mdi-supp.pdf. We thank Isabelle Guyon for the idea for the paper and Joan Bruna and seminar participants at the University of California, Berkeley for comments. Poulos acknowledges support from the National Science Foundation Graduate Research Fellowship [grant number DGE 1106400].

[‡]Corresponding author mailing address: Department of Political Science, 210 Barrows Hall #1950, Berkeley, CA 94720-1950. E-mail: poulos@berkeley.edu.

## 1. Introduction

Supervised learning has become an increasingly attractive methodology and proven to be effective in social science applications, such as studies of international and civil conflict [2, 6, 15] and election fraud [3, 14]. For supervised classification tasks, the objective is to fit a model on labeled training data in order to categorize new examples. However, the ability of researchers to accurately fit a model and yield unbiased estimates may be compromised by missing data.

Our objective is to compare the out-of-sample performance of three popular machine learning classifiers — decision trees, random forests, and neural networks — trained on imputed or non-imputed (i.e., one-hot encoded) machine learning benchmark datasets that contain various degrees of missing-data perturbation. Researchers analyzing survey data typically choose decision trees or random forests for classification tasks, largely because these models do not require imputing missing data nor encoding categorical variables, unlike neural networks or other classifiers.

The primary contribution of this paper is to provide guidance to applied researchers on how to handle missing data for supervised learning tasks. First, we show that imputation methods can increase predictive accuracy in the presence of missing-data perturbation. Second, we show that adding missing-data perturbation prior to imputation can actually improve prediction accuracy by regularizing the classifier. We achieve the state-of-the-art on the Adult dataset with missing-data perturbation and $k$-nearest-neighbors ($k$-NN) imputation. Lastly, we show that classifiers trained on one-hot encoded data generally yield higher predictive accuracy when the data are not additionally perturbed. For example, a simple one-hot encoded random forests outperforms the state-of-the-art on the Congressional Voting Records (CVRs) dataset with no missing-data perturbation.

This manuscript is organized as follows: Section 2 describes missing data mechanisms and imputation methods; Section 3 describes our experiments on two benchmark datasets and discusses the results; Section 4 concludes and offers possibilities for future research.

## 2. Missing data and imputation methods

In this section, we describe the missing data mechanisms underlying patterns of missing data common to survey-based datasets. We then review popular methods of handling missing data.

### 2.1. Missing data patterns and mechanisms

It is important to first distinguish between missing data patterns, which describe observed and missing values, and missing data mechanisms, which relate the probability of missingness. [12, Chap. 1]. Common missing data patterns in surveys typically include unit nonresponse, where a subset of participants do not complete the survey, and item nonresponse, where missing values are concentrated on particular questions. In opinion polls, nonresponse may reflect either refusal to reveal a preference or lack of a preference [4].

Following the notation of Little and Rubin [12], let $Y = y_{ij}$ be a $(n \times K)$ dataset with each row $y_i = (y_{i1}, \ldots, y_{iK})$ the set of $y_{ij}$ values of feature $Y_j$ for example $i$. Let $Y_{\text{obs}}$ define observed values of $Y$ and $Y_{\text{mis}}$ define missing values. Define the missing data identity matrix $M = m_{ij}$, where $m_{ij} = 1$ if $y_{ij}$ is missing and $m_{ij} = 0$ if $y_{ij}$ is nonmissing. The missing data mechanism is missing completely at random (MCAR) if the probability of missingness is independent of the data, or

$$f(M|Y, \phi) = f(M|\phi) \quad \forall Y, \phi,$$

where $\phi$ denotes unknown parameters. The missing at random (MAR) assumption is less restrictive than MCAR in that the probability of missingness depends only on the observed data, $f(M|Y,\phi) = f(M|Y_{\text{obs}},\phi)$ for all $Y_{\text{mis}}, \phi$. The missing not at random (MNAR) assumption is that the probability of missingness may also depend on the unobserved data, $f(M|Y,\phi) = f(M|Y_{\text{mis}},\phi)$ for all $Y_{\text{mis}}, \phi$. Researchers typically assume data is MAR, which mitigates the identifiability problems of MNAR because the probability of missingness depends on data that are observed on all individuals [21, Chap. 6].

## 2.2. Imputation methods

Complete-case analysis (i.e., discarding examples with missing values) wastes information and biases estimates unless the missing data are MCAR. Since there is no way to distinguish whether the missing data are MCAR or MNAR from the observed data, a natural strategy is to impute missing values and then proceed as if the imputed values are true values. Imputation methods that rely on explicit model assumptions include *mean or mode replacement*, which substitutes missing values with the mean (for quantitative features) or mode (for qualitative features) of the feature vector, and *prediction model* imputation, which replaces missing values with the predicted values from a regression of $Y_{\text{mis}}$ on $Y_{\text{obs}}$.

Explicit modeling methods assume the data are MAR while implicit modeling methods, which are algorithmic in nature and rely only on implicit assumptions, generally do not assume the underlying missing data mechanism. Implicit methods include *random replacement*, where an example with missing data is randomly replaced with another complete example randomly sampled, and *hot deck* imputation, where missing values are replaced by "similar" nonmissing values. Hot deck imputation can be implemented by computing the $k$-nearest-neighbors ($k$-NN) of an example with missing data and assigning the mode of the $k$-neighbors to the missing data. Batista and Monard [1] use this procedure and find $k$-NN imputation can outperform summary statistic imputation and internal methods used by decision trees to treat missing data.[3]

---

[3]Li et al. [10] propose a hot deck imputation method based on fuzzy $k$-means.

In related work, Silva et al. [19] empirically compare imputation using neural networks with mean/mode imputation, regression models (logistic regression and multiple linear regression), and hot deck, finding neural networks performs the best on datasets with categorical variables.

### 2.3. One-hot encoding

Another natural strategy in dealing with missing data for supervised learning problems is one-hot encoding. Instead of imputing missing data, one-hot encoding creates a binary feature vector that indicates missing values. For categorical features, one-hot encoding simply treats a missing value symbol (e.g, "?") as a category when the categorical features are binarized. For continuous features, missing values are set to a constant value and a missingness indicator is added to the feature space. One-hot encoding for missing data yields biased estimates when the features are correlated, which is often the case with survey data, even when data are MCAR [8].

## 3. Experiments

In this section, we describe our experiment on two machine learning benchmark datasets with missing categorical data, comparing three popular classifiers — neural networks, decision trees, and random forests— trained on either one-hot encoded or imputed data with different degrees of MCAR perturbation.

### 3.1. Benchmark datasets

We experiment on two benchmark datasets from the UCI Machine Learning Repository: the Adult dataset and CVRs dataset [11]. The Adult dataset contains $N = 48,842$ examples and 14 features (6 continuous and 8 categorical). The prediction task is to determine whether a person makes over $50,000 a year. The CVRs dataset contains $N = 435$ examples, each

the voting record of a member of the U.S. House of Representatives for 16 key roll call votes. The dataset contains 16 categorical features with three possible values: "yea", "nay", and missing. The prediction task is to classify party affiliation (Republican or Democrat). In contrast to the Adult dataset, in which only a few features are highly correlated, many of the roll call votes in the CVRs dataset exhibit strong correlations (Figures SM-1 and SM-2).

The state–of–the–art for the Adult dataset is a Naive Bayes classifier that achieves a 14.05% generalization error after removing examples with missing values [9]. The CVRs dataset donor claims to achieve a 5-10% error rate using an incremental decision tree algorithm called STAGGER, although it is unknown to the authors what train-test split is used or how missing values are handled [17, 18].

### 3.2. Patterns of missing data

Uncovering missing data patterns in the datasets will help to identify possible missing data mechanisms and select appropriate imputation methods. Figure SM-3 analyzes patterns of missing data in the Adult dataset, in which 7% of the examples contain missing values. Missing data in the Adult dataset is due to item nonresponse, as missing values are concentrated in three of the categorical features — *Work class*, *Occupation*, and *Native country*— and no examples contain entirely missing data. It is unlikely that the data are MCAR because observations that are missing in *Work class* are also missing in *Occupation* (about 6% of examples have missing values in both).

Missing values in the CVRs dataset are not simply unknown, but represent values other up-or-down votes, such as voted present, voted present to avoid conflict of interest, and did not vote or otherwise make a position known. Close to half of the CVRs data contains missing values, which are present in every feature (Figure SM-4). About a quarter of missing data is in `South Africa`, which was a controversial amendment to amend the Export Administration Act to bar U.S. exports to South Africa's apartheid regime. Twelve percent of missing data is in the feature `Water`, which is a water projects authorizations bill, and 7% of missing data rests in the feature `Exports`, which is a tariff bill. The data are unlikely to be MCAR because 12% of the data are missing in just `South Africa` and less than 1% of examples are missing across all features. It is most likely in this case that the CVRs data are MNAR because the probability of missing a vote or voting present on one important bill should not theoretically be influenced by observed votes on other important bills.

### 3.3. Preprocessing

After randomly splitting each dataset 2/3 for training and 1/3 for testing, we perturb the training data so that the proportion of missing values in the set of categorical features $Y_{\text{cat}}$ follows $\delta = \{0.1, 0.2, 0.3, 0.4\}$ according to the MCAR mechanism

$$\Pr(M = 1 | Y_{\text{cat}}, \phi) = \delta \text{ for all } Y_{\text{cat}}.$$

We use missing-data perturbation to study the impact of larger amounts of missing data; however, it is also a form of dropout noise that can be used to control overfitting during the training process and improve the generalizability of the model [22].

After one-hot encoding the categorical variables in the training data, we implement each of the following imputation techniques, discussed in Section 2.2: $k$-NN, prediction model (logistic regression, random forests, or SVMs), mode replacement, and random replacement. We then standardize continuous features by subtracting the mean and dividing by the standard deviation of the feature. The test data is preprocessed in the same manner, with the exception that we do not perturb categorical features in the test data.[4]

### 3.4. Model training

---

[4]When imputing the missing data with mode replacement, we use the training set mode. We also use the training set mean and standard deviation to standardize test set features.

We train three different classifiers on the preprocessed data: decision trees, random forests, and neural networks. The neural networks consist of four layers, each of the two hidden layers having 1024 nodes, and updates with the adaptive learning rate method *Adadelta* [24]. We explore the hyperparameter space — momentum schedule, dropout regularization, and learning rate — using Bayesian optimization [20], which selects optimal models using the mean training error rate as our objective function. Figure SM-5 shows the exploration of hyperparameter space during Bayesian optimization for both datasets. Random forests and decision trees are trained with preselected hyperparameters.

### 3.5. Results

We assess the performance of the classifiers in terms of test set error rate on one-hot encoded or imputed data and for various levels of MCAR perturbation. The results on the Adult dataset and CVRs dataset are plotted in Figures 1 and 2, respectively, with error bars representing $\pm 1$ standard deviation from the test error rate.

One-hot-encoded decision trees outperforms the state-of-the-art on the CVRs dataset by over 2% ($0.027 \pm 0.006$). The neural networks classifier trained on data imputed with $k$-NN yields the lowest generalization error ($0.144 \pm 0.06$) on the Adult dataset with 10% of the categorical feature values perturbed, which is comparable to the state-of-the-art. In comparison, a random forests classifier trained on non-perturbed and one-hot encoded data yields a test error rate of $0.152 \pm 0.02$. This comparison shows that the classifiers can overfit the data and, in the case of imputed models, perturbation improves prediction accuracy by regularizing the classifier.

Overall, the results show imputation methods can increase predictive accuracy in the presence of missing-data perturbation. For both datasets, one-hot encoded models trained in the absence of perturbation perform as well as imputed models trained on non-perturbed data. In the case of the Adult dataset, imputation clearly improves accuracy in the presence of MCAR-perturbed data. In contrast, each of the three classifiers trained on the one-hot encoded CVRs dataset perform relatively well across different levels of perturbation. The general pattern of results hold when the classifiers are trained on MNAR-perturbed data (Figures SM-6 and SM-7).

## 4. Conclusion

This paper compares methods for imputing missing categorical data for supervised classification tasks. We compare the out-of-sample performance of neural networks against decision tree and random forest classifiers trained on datasets with one-hot encoded or imputed data, across different levels of MCAR-perturbed training data. Our results are comparable to the state-of-the-art on the Adult dataset using a neural networks classifier, $k$-NN imputation, and MCAR data-perturbation. $k$-NN imputation likely performs well in this case because it is an implicit modeling method that does not assume the underlying missing data mechanism. This result is in line with Batista and Monard [1], who find $k$-NN imputation can outperform explicit modeling methods for supervised learning tasks.

We conclude from the results that the performance of the classifiers and imputation strategies generally depend on the nature and proportion of missing data. For the Adult dataset, neural networks trained on imputed data generally outperform other classifiers and imputation methods across different ratios of perturbed data, while classifiers trained on one-hot encoded data perform very poorly on perturbed training data.

The results of the present study show that perturbation can help increase predictive accuracy for imputed models, but not one-hot encoded models. Future work can identify the conditions under which missing-data perturbation can improve prediction accuracy. Interesting extensions of this paper include evaluating the benefits of using missing-data perturbation over more popular regularization techniques such as dropout training [7, 13, 23].

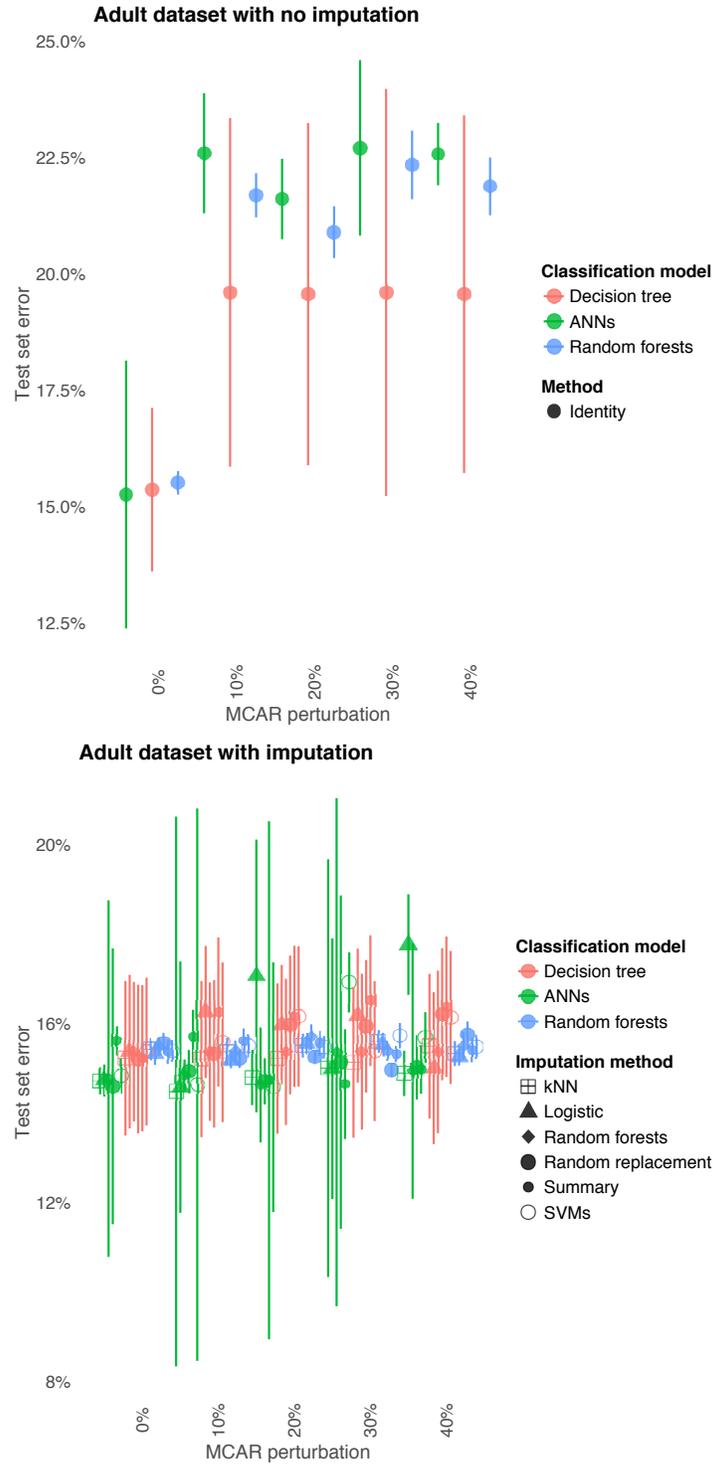

Figure 1: Error rates on the Adult test set with (bottom) and without (top) missing data imputation, for various levels of MCAR-perturbed categorical training features (x-axis). For neural networks, prediction intervals are obtained from the standard deviation of test set errors of neural networks trained with different convergences [5]. For random forests and decision trees, prediction intervals follow from the variation created by varying the maximum depth of the decision trees, and for random forests, the number of trees and decision rule for the number of features to consider when looking for the best split.

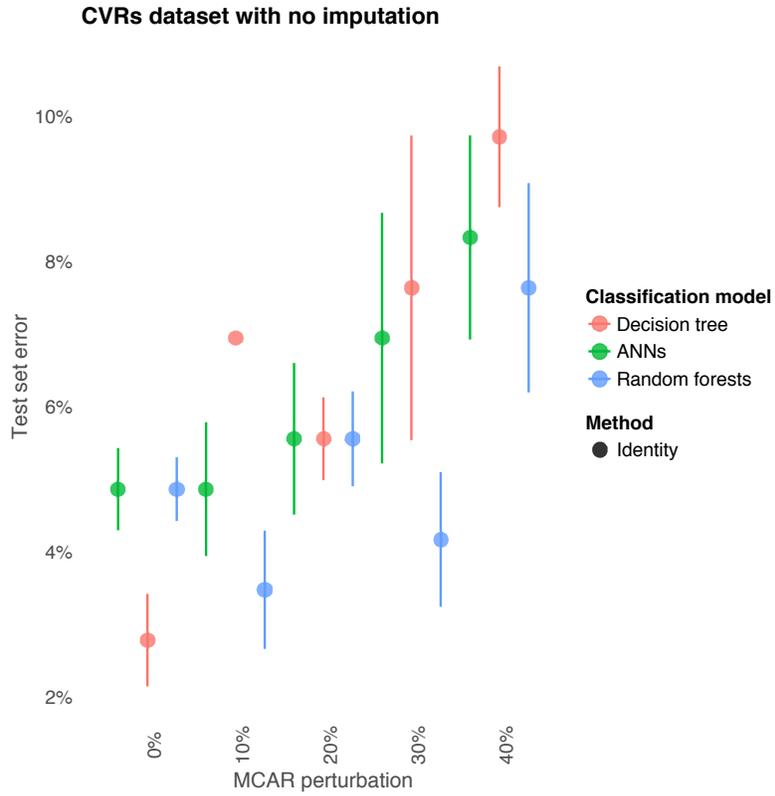

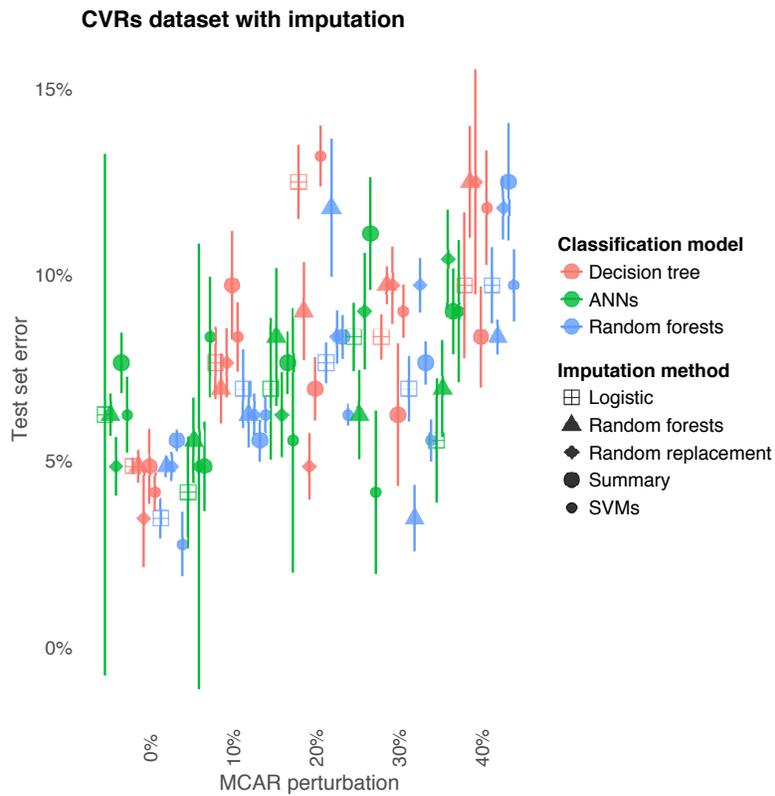

Figure 2: Error rates on the CVRs test set with (bottom) and without (top) missing data imputation. See footnotes for Figure 1.